# Halving transcription time: A fast, user-friendly and GDPR-compliant workflow to create AI-assisted transcripts for content analysis

**Jakob Sponholz**[1]
University of Cologne,
Germany
jakob.sponholz@uni-koeln.de

**Andreas Weilinghoff**[2]
University of Koblenz,
Germany
weilinghoff@uni-koblenz.de

**Juliane Schopf**[3]
University of Münster,
Germany
juliane.schopf@uni-muenster.de

## Abstract

In qualitative research, data transcription is often labor-intensive and time-consuming. To expedite this process, a workflow utilizing artificial intelligence (AI) was developed. This workflow not only enhances transcription speed but also addresses the issue of AI-generated transcripts often lacking compatibility with standard content analysis software. Within this workflow, automatic speech recognition is employed to create initial transcripts from audio recordings, which are then formatted to be compatible with content analysis software such as ATLAS or MAXQDA. Empirical data from a study of 12 interviews suggests that this workflow can reduce transcription time by up to 76.4%. Furthermore, by using widely used standard software, this process is suitable for both students and researchers while also being adaptable to a variety of learning, teaching, and research environments. It is also particularly beneficial for non-native speakers. In addition, the workflow is GDPR-compliant and facilitates local, offline transcript generation, which is crucial when dealing with sensitive data.

[1] Jakob Sponholz (Corresponding Author)
https://orcid.org/0000-0002-2555-0565
[2] Andreas Weilinghoff
https://orcid.org/0009-0005-3754-8285
[3] Juliane Schopf
https://orcid.org/0000-0002-7527-7464

## 1 Introduction

The transcription of spoken language is an essential part of numerous research, teaching and study projects, both within and beyond academia. Working with transcripts is widespread across various fields, including linguistics, social and educational sciences, psychology, and journalism, as well as legal contexts such as court hearings and police interrogations (cf. Schmidt, 2023, p. 13). The requirements for transcription, i.e. the medial transition from orality to writing (cf. Deppermann, 1999, p. 39), are highly heterogeneous: in the field of linguistic conversation analysis, for instance, the aim is to represent as many typical spoken language phenomena as possible in writing (utterance breaks, parallel speech, pitch changes, accents, phonetic variance in pronunciation, etc.) (cf. Imo & Lanwer, 2019). In this and other contexts, the process of transcription already constitutes an essential component of the scientific method (cf. Schmidt, 2023, p. 24) itself.

The conventions according to which such transcripts are made include numerous special characters and deviations from standard orthography (e.g., "·" in HIAT for pauses or ";" in GAT 2 for a mid-rising pitch movement at the end of an intonation phrase). Therefore, transcribers must first master the transcription rules before they can make appropriate transcripts. However, the transcription task involves more than just the transformation of speech into text. It is also part of the analysis because it requires an understanding for the conversation, the participant's perspective and experience-based knowledge which can only be accomplished by human transcribers (cf. Deppermann, 2007, p. 27). Due to these sophisticated skills, it is currently difficult to imagine that transcripts can be fully produced by AI systems in conversation analysis.

The use of transcriptions varies across disciplines. In the social sciences, for example, interview transcriptions for qualitative content analysis (cf. Braun & Clarke, 2006; Kuckartz & Rädiker, 2024; Mayring, 2021) typically require less detail. In this type of transcription, the peculiarities of oral conversations are disregarded and the recording is transferred into standard orthographic written language. As a result, this form of transcription is not part of the hermeneutic procedure but serves as necessary preparatory work aligned with established spelling conventions. Since the primary focus in this context is on content rather than form, AI-based transcription is a well-suited solution for this. This article deals with this type of transcription.

Currently, a wide range of options for automated transcription exist, ranging from commercial online services that charge on an hourly basis or via minute-based quotas to complete software packages. However, the use of most of these services is problematic in an academic research context: Data protection regulations and ethics committees of most universities require researchers to handle data in compliance with General Data Protection Regulation (GDPR). This means, that audio or video data must not be processed online or uploaded to servers or clouds hosted by companies outside the EU. In addition, most scientific projects lack the financial means to purchase full licenses for transcription software. Moreover, there are complaints about the difficulty of seamlessly integrating automatically generated transcripts with content analysis software. This lack of compatibility poses a challenge for researchers and analysts who rely on successful integration for effective content analysis. Addressing these compatibility issues is essential to improving the usefulness and efficiency of automated transcription services across various professional settings.

Thus, there is a need for free, unrestricted automatic transcription that can be performed on local computers and offline, with resulting transcripts that are fully compatible with appropriate content analysis software. This paper presents a 3-step workflow to achieve this goal. Using the audio transcription feature in Adobe® Premiere Pro® CC[4], this workflow creates an AI-generated pre-transcript that can be transferred to content analysis software. This workflow includes the following three steps:

1. Adobe ® Premiere Pro® CC is used to create a pre-transcript. A free trial version of the program is available.
2. Either Microsoft Word or a GDPR-compliant Web-App is used to prepare the AI-assisted pre-transcript from Adobe® Premiere Pro® CC for import into a content analysis software.
3. Content Analysis Software is used to revise, finalize and analyze the transcript.

During the process, a pre-transcript is generated using the following timestamp format: [hh:mm:ss.xx]. This format can be read by multiple content analysis softwares, e.g. ATLAS.ti (Scientific Software Development GmbH, 2025) and MAXQDA (VERBI Software Consult Sozialforschung GmbH, 2025). The alignment of the timestamps with the audio file helps the researcher to jump from a text passage to the corresponding audio passage during analysis. An additional advantage of this workflow is that transcription can take place offline and locally on the hard drive. According to Adobe®, using Premiere Pro® CC (Version 22.2 or higher) for speech to text applications is GDPR-compliant (cf. Adobe, 2025). This workflow for creating verbatim transcripts of audio or video files can be useful in a variety of contexts and usage scenarios. Furthermore, it supports not only the transcription of audio tracks with a single speaker (e.g. voice memos), but also, through its speaker recognition function, the processing of conversations with two or more participants (e.g. dialogues, interviews, group discussions).

[4] This workflow is presented as an example; there are a variety of other automatic transcription software, but most of them charge on an hourly basis. Other tools for creating AI-generated pre-transcripts include, for example, Whisper, Buzzcaptions, JupyterHub and Verbatim Transcripts. It should be noted, however, that not all of these tools (1) can be used offline, (2) are user-friendly and (3) are free of charge. This workflow using Adobe® Premiere Pro® provides a (at least for the trial period of 7 days) free opportunity to convert an unlimited amount of audio- and video recorded conversations into written transcripts offline.

The audio transcription feature in Adobe® Premiere Pro® CC supports a wide range of languages. These include English, Chinese (simplified and traditional; Mandarin), Spanish, German, French, Japanese, Portuguese (European) and Italian. It also supports Korean, Russian, Hindi, Dutch, Norwegian, Swedish, and Danish. These transcripts can also improve accessibility in inclusive settings – especially when using audio or videos files in educational contexts. In the following, we will first explain the workflow step by step (see Section 2) and apply it for a specific interview (see Section 3). Finally, in Section 4 the limitations are discussed.

## 2 Workflow and implementation

This section describes the workflow step by step by executing the three parts of the procedure: creating the pre-transcripts (2.1), preparing the pre-transcripts for content analysis software (2.2) and importing the pre-transcripts into the content analysis software (2.3). The workflow requires Adobe® Premiere Pro® CC (Version 22.2 or higher) and a text editor or data analysis software of your choice capable of reading the provided timestamp formatting. For data preparation, users can choose either Microsoft Word or the fully automated Web-App-Version (see 2.2).

### 2.1 Creating the pre-transcript

To create the pre-transcript, first prepare a project file in Premiere Pro®. In this project file, add a separate sequence for each file to be transcribed. To illustrate the respective steps described below, we use a conversation from the freely accessible "Scottish Corpus of Texts & Speech", namely a conversation between two students (The Scottish Corpus of Texts & Speech, 2025). This also tests how the AI deals with different language varieties, such as Scottish English in this example.

Be sure to name each sequence accordingly so that you can easily associate the sequence with the file. After that, please perform the following steps for each sequence:

1. Start a new project in Premiere Pro®. Create a new sequence. Insert your audio or video file.
2. Access Auto-transcription by clicking *Window/Text*. Open the Transcription-Tab.
3. Select '…' at the top of the window to choose your language: Enable the option *recognize when different speakers are talking*. Click *Transcribe*.
4. Find your sequence in the library. Double-click to open your sequence in the timeline.
5. In the Transcript tab, click on '…' at the top of the window to access export options.

6. Export your transcript by clicking *Export to text*. This is your pre-transcript. *Note: Make sure you export your text file from Transcript (and not from Captions).*

7. Save your pre-transcript.

## 2.2 Preparation of the pre-transcripts for analysis software

To prepare the pre-transcripts for analysis software, we propose two options. The first option implements the use of Microsoft Word and does not require a high level of technical expertise. Hence, the first option is particularly user-friendly and easily accessible because it implements a well-known text editing software. The second option uses a GDPR-compliant Web-App that was developed to speed up the transformation of the speaker labels. Using this Web-App significantly speeds up the preparation process. Depending on the project requirements, users may need to modify or adapt the speaker labels in the transcript. Accordingly, we provide two methods: one that keeps the original speaker labels and another that adapts or removes them. This task can be performed either in Microsoft Word or via the GDPR-compliant Web-App, which was designed solely for this purpose. In summary, users can choose to implement either Word or the Web-App and decide whether to retain or adapt speaker labels. All four options ensure that the pre-transcripts are properly formatted for analysis software. Hence, users can select the most suitable approach based on their project goals and personal technical skills.

### 2.2.1 Option A: Keep speaker labels

This workflow uses Microsoft Word to prepare the pre-transcripts and keep the speaker labels. The speaker labels from Adobe® Premiere Pro® CC are automatically generated in the format "Speaker 1" for English settings, but this may be adapted according to the user language. The following commands adapt the timestamps so that the transcript can be imported to the analysis software.

1. **Open Microsoft Word and create a new document.**

2. **Copy and paste the text of your pre-transcript into your document.**

   **Example of a pre-transcript before formatting:**

```
00:02:01:16 - 00:02:07:09
Speaker 1
I think if you're in a totally neutral environment where you actually have no
responsibility and I think that you get on better.

00:02:07:11 - 00:02:18:22
Speaker 2
Yeah, sometimes I think, well, sometimes I think it's good when I get home. It's
quieter in the sense of if I was a house residence last year, this year with my
friends and they're in flats, think that's right.
```

3. Open the **"Search and replace"**-feature. Go to **"Replace"**.

4. Go to **"Advanced find and replace"**.
   *Note: The path to this feature varies depending on the operating system and the version deployed.*

5. **Activate "Use wildcard"** in the Options-checkbox.

6. **Change all timestamps:**

   **Find:**
   ```
   (??):(??):(??):(??)
   ```

   **Replace with:**
   ```
   [\1:\2:\3.\4]
   ```

   **Click "Replace all".**
   *Note: Make sure the document you are editing is selected.*

   **Result:**
   ```
   [00:02:01.16] - [00:02:07.09]
   Speaker 1
   I think if you're in a totally neutral environment where you actually have no
   responsibility and I think that you get on better.

   [00:02:07.11] - [00:02:18.22]
   Speaker 2
   Yeah, sometimes I think, well, sometimes I think it's good when I get home. It's
   quieter in the sense of if I was a house residence last year, this year with my
   friends and they're in flats, think that's right.
   ```

7. **Remove second timestamp:** Find the following expression and leave the "Replace with"-field empty.

   **Find:**
   ```
   ( - [[])(??):(??):(??).(??)([]])
   ```

   **Replace with:**
   ```
   (nothing, leave it empty)
   ```

   **Click "Replace all”.**

8. **Result:**
   ```
   [00:02:01.16]
   Speaker 1
   I think if you're in a totally neutral environment where you actually have no
   responsibility and I think that you get on better.
   ```

9. **Deactivate "Use wildcard".**

The resulting pre-transcript is now ready to imported into a content analysis software or to be further modified (see chapter 2.2.2)

### 2.2.2 Option B: Change or remove speaker labels

After the adaptation of the timestamps (see chapter 2.2.1), the following commands allow the user to adapt or remove the speaker labels via Microsoft Word.

**Remove "Speaker 1"** and **replace** it with a different designation (e.g. name, role or pseudonym). If you want to remove the speaker, replace "Speaker 1" with a space key:

1. Open the **"Search and replace"**-feature. Go to **"Replace".**

2. Go to "**Advanced find and replace"**.

   **Find:**
   `Speaker 1`

   **Replace with:**
   `Designation of your speaker (e.g. pseudonym, interviewer, name of the speaker). If you want to remove the speaker, replace with a space key. In this example, the women speaking are called "Bonnie" and "Fiona".`

   **Click "Replace all"**.
   *Note: Repeat step with all speakers (Speaker 2, Speaker 3...)*

This would be an example for the exchange of the speaker labels:

3. **Find:**
   `^pSpeaker 1^p`

   **Replace with:**
   `Bonnie:`

   **Result:**
   `[00:02:01.16] Bonnie: I think if you're in a totally neutral environment where you actually have no responsibility and I think that you get on better.`

This would be an example for the deletion of the speaker labels:

4. **Find:**
   `^pSpeaker 1^p`

   **Replace with:**
   `[Space key]`

   **Result:**
   `[00:02:01.16] I think if you're in a totally neutral environment where you actually have no responsibility and I think that you get on better.`

5. **Select all text in this document and save this text into a .txt-file.**

### 2.2.3 Automating this workflow: The Transcript Timestamp Wizard

Based on the workflow described, a GDPR-compliant web app was developed. The "Transcript Timestamp Wizard" (Sponholz & Müller, 2024a) automates the steps previously performed manually in Microsoft Word. No prior knowledge is required – users do not need to complete the individual steps themselves, making the process significantly simpler and faster. The open-source app is available on GitHub (cf. Sponholz & Müller, 2024b) under the MIT license and can be used and redistributed free of charge. As a result, this application contributes to the sustainable use of artificial intelligence at universities, benefiting the scientific community, research-led teaching, and aiding students alike.

## 2.3 Import of pre-transcripts into the content analysis software

Your AI-generated, verbatim transcript is now ready to be imported to content analysis software. During analysis, the aligned timestamps will allow you to jump right from a paragraph in the transcript to the corresponding passage in the audio file.

# 3 An applied example from empirical research

In the following, we provide an empirical testing of this workflow. For this case study, we used data from a PhD project in rehabilitation sciences, consisting of 12 interviews from the year 2023. All interviews together have a total length of 09:40:49 hours and last between 21:49 minutes and 01:12:08 hours each. The data represent dyadic interactions involving the doctoral student and individual participants. The interviews were guided, semi-structured conversations conducted in German, with all participants hailing from North Rhine-Westphalia and speaking standard German. As the doctoral student's research focuses on the barriers and resources of students with disabilities and chronic illnesses, all 12 participants (3 male, 9 female) have a physical disability or chronic illness and some also have mental illnesses and/or visual impairments. For two participants (Interview IDs 1 and 12), muscular atrophy affects their speech, resulting in slurred articulation. Given these factors, this corpus is of particular interest to us for examining how artificial intelligence handles these difficult transcription conditions. The transcription of the data in Premiere Pro (version 24.0.0) took 23:52 minutes on a Mac Studio (Mac 13,1; Chip: Apple M1 Max; 10-Core CPU; 24-Core GPU; 32 GB LPDDR5 RAM). Adobe® Premiere Pro® CC provides the option to generate a .csv file, a .txt file and a .prtranscript file as transcription outputs. While the automatically generated pre-transcripts require manual correction, this process is significantly more efficient compared to starting a transcription from scratch. For instance, a manual transcription of a one-hour interview typically takes between 5 to 10 hours (Kuckartz & Rädiker, 2024, p. 203). In this study, the doctoral student spent a total of 22 hours and 21 minutes manually correcting all 12 interviews. Over 19 work sessions, the student transcribed and corrected approximately 9 hours and 40 minutes of audio, averaging 2 hours and 19 minutes per hour of interview. As a result, this workflow substantially reduced transcription time, with savings ranging from 53.8% to 76,9% compared to traditional manual transcription. Beyond the notable time savings, there is an additional analytical advantage, as the transcripts are corrected directly within the MAXQDA software, initiating text work during the transcript check itself (Kuckartz & Rädiker, 2024, p. 203). Based on the AI-generated pre-transcripts and the corrected transcripts, the word error rate (WER) could then be calculated to quantify transcription accuracy. The WER indicates how many words - as a percentage of the total - were incorrectly recognized. Based on all transcripts, the WER amounted to 17.4% in our case study. Table 1 provides an overview of the working times, interview lengths and word error rates:

**Table 1**

*Transcription times and word error rates of data sample*

| Interview ID | Work time | Interview duration | Work time / Interview duration | Interview duration / work time | Word error rate |
|---|---|---|---|---|---|
| 1 | 01:42:00 | 00:34:40 | 2.942 | 0.340 | 0.1847 |
| 2 | 02:02:00 | 00:44:26 | 2.746 | 0.364 | 0.1990 |
| 3 | 01:23:00 | 00:44:07 | 1.881 | 0.532 | 0.1604 |
| 4 | 02:27:00 | 00:49:10 | 2.990 | 0.334 | 0.1924 |
| 5 | 01:19:00 | 00:27:14 | 2.901 | 0.345 | 0.1961 |
| 6 | 02:37:00 | 01:03:07 | 2.487 | 0.402 | 0.1620 |
| 7 | 02:33:00 | 01:12:08 | 2.121 | 0.471 | 0.1552 |
| 8 | 01:38:00 | 00:51:56 | 1.887 | 0.530 | 0.1535 |
| 9 | 01:26:00 | 00:44:50 | 1.918 | 0.521 | 0.1628 |
| 10 | 00:58:00 | 00:34:09 | 1.698 | 0.589 | 0.1872 |
| 11 | 01:33:00 | 00:51:54 | 1.792 | 0.558 | 0.1498 |
| 12 | 02:43:00 | 01:03:08 | 2.582 | 0.387 | 0.1798 |
| Sum/mean: | 22:21:00 | 09:40:49 | 2.309 | 0.433 | 0.1736 |

With regard to the question of which linguistic and non-linguistic factors caused errors in speech recognition, the following three factors, which are known in the literature (Jurafsky & Martin, 2025, p. 333) were particularly decisive:

- Background noise: speech recognition was not always able to distinguish between the speech of the interviewees and background noise. In some cases, background noises, such as the movement of a joystick on a wheelchair, were interpreted as speech material and were usually labeled with hesitation markers such as "uh" or "hm".

- Fast speaking speed and technical language: In some cases, fast speaking led to incorrect transcription as not all lexemes could be recognized and successfully separated. This was particularly noticeable with technical terminology, for example, when the word "Barrieren" (engl. "barriers") was processed as "Bayern" (engl. "Bavaria").

- Changes of origo: Deictic changes of perspective, such as those triggered by speech renditions or animated speech, sometimes caused difficulties for automatic speech recognition. For example, speech renditions that are integrated into one's own utterances are separated and provided with incorrect pronouns.

The above-mentioned question of whether utterances from individuals with speech difficulties due to muscle atrophy pose challenges for automatic transcription can be clearly answered in the negative: The word error rate for these two interviews (IDs 1 and 12) was not higher than in the other interviews. This suggests that the workflow is also suitable for transcribing speech data from individuals whose articulation may be less clear due to disabilities, age, or illness. However, as expected, automatic speech recognition struggled more with background noise and very fast speech (Jurafsky & Martin, 2025, p. 333), requiring additional corrections. For data collection, this implies that attention should be paid to environments with as little background noise as possible and that more time should be allowed for the transcription of fast-speaking interviewees, as is also the case with manual transcription.

## 4 Discussion

This paper presents a fast, user-friendly, and GDPR-compliant workflow for creating AI-assisted transcriptions that are suitable for content analysis. By using Adobe® Premiere Pro® CC for generating pre-transcripts, supplemented by tools like Transcript Timestamp Wizard for formatting, this method allows the transcription time to be halved compared to manual transcription. This approach can benefit a diverse range of users, including students, academics, non-native speakers, and individuals with hearing impairments, by enhancing accessibility and transcription accuracy. This method allows analysis of various types of spoken data, such as interviews and group discussions. Additional features include GDPR compliance and the option to select from a wide range of languages. It should also be noted that multilingual data cannot be processed within just one execution of this workflow. While Adobe® Premiere Pro® CC can transcribe different languages, it requires users to select a single language per transcription. Although language switching is not possible within a single transcription, each language can be transcribed separately and then combined after that. Furthermore, the method allows linking text segments to their corresponding audio or video files within the corresponding analysis software.

For a critical reflection on the workflow and as a starting point for further research, the following limitations should be taken into account: The transcription is provided in standard orthography, making this workflow particularly useful for content analysis applications. Analyses of linguistic variation and changes in variety, allegro forms, cliticization, listeners' responses, inhalations/exhalations, laughter, pauses, and prosodic features cannot be made using this AI-powered transcription method. Furthermore, it should be noted that the quality of the transcription depends on the voice recording. The reported halving of the transcription time applies to optimal recording conditions and varieties close to the standard language. In cases of loud background noise, a large number of speakers, overlapping speech, regional variants, or strong accents, automatic speech recognition makes a more intensive manual correction necessary.